\documentclass[11pt, a4paper, twoside, openany]{article}
\input epsf
\usepackage{t1enc}
\usepackage{color}
\usepackage{epsfig}
\usepackage{pictex}
\usepackage{ifthen}
\usepackage{fancyheadings}
\usepackage{natbib}
\newcommand{\psfigureab}[5]{
        \begin{figure}[h!]
        \smallskip
        \begin{center}
	\begin{tabular}{cc}
	        \epsfxsize=#4 \epsfbox{#2.eps} &
        	\epsfxsize=#4 \epsfbox{#3.eps} \\
		(a) & (b) \\
	\end{tabular}
        \end{center}
        \caption{#5}
        \label{fig:#1}
        \end{figure}
}

\newcommand{\psfigureabc}[6]{
        \begin{figure}[h!]
        \smallskip
        \begin{center}
	\begin{tabular}{ccc}
	        \epsfxsize=#5 \epsfbox{#2.eps} &
        	\epsfxsize=#5 \epsfbox{#3.eps} &
        	\epsfxsize=#5 \epsfbox{#4.eps} \\
		(a) & (b) & (c) \\
	\end{tabular}
        \end{center}
        \caption{#6}
        \label{fig:#1}
        \end{figure}
}

\newcommand{\psfigureabcdef}[9]{
        \begin{figure}[h!]
        \smallskip
        \begin{center}
	\begin{tabular}{cc}
	        \epsfxsize=#8 \epsfbox{#2.eps} &
        	\epsfxsize=#8 \epsfbox{#3.eps} \\
		(a) & (b) \\[14pt]
	        \epsfxsize=#8 \epsfbox{#4.eps} &
        	\epsfxsize=#8 \epsfbox{#5.eps} \\
		(c) & (d) \\[14pt]
	        \epsfxsize=#8 \epsfbox{#6.eps} &
        	\epsfxsize=#8 \epsfbox{#7.eps} \\
		(e) & (f) \\
	\end{tabular}
        \end{center}
        \caption{#9}
        \label{fig:#1}
        \end{figure}
}
\newcommand{\psfigureabcdefhoriz}[9]{
        \begin{figure}[h!]
        \smallskip
        \begin{center}
	\begin{tabular}{ccc}
	        \epsfxsize=#8 \epsfbox{#2.eps} &
        	\epsfxsize=#8 \epsfbox{#3.eps} &
        	\epsfxsize=#8 \epsfbox{#4.eps} \\
		(a) & (b) & (c) \\[14pt]
	        \epsfxsize=#8 \epsfbox{#5.eps} &
        	\epsfxsize=#8 \epsfbox{#6.eps} &
        	\epsfxsize=#8 \epsfbox{#7.eps} \\
		(d) & (e) & (f) \\
	\end{tabular}
        \end{center}
        \caption{#9}
        \label{fig:#1}
        \end{figure}
}

\begin{document}
\begin{center}
	\Large
	Feedforward Neural Networks with \\
	Diffused Nonlinear Weight Functions \\

\vspace{0.2in}

	\large
	Artur Rataj, e--mail arataj@iitis.gliwice.pl \\
	Institute of Theoretical and Applied Computer Science \\
	of the Polish Academy of Sciences, Ba\l tycka 5, Gliwice, Poland

\vspace{0.2in}

	\large
	Technical Report IITiS-2003-02-23-1-1.06

\end{center}
\vspace{0.1in}

\begin{abstract}
In this paper, feedforward neural networks are presented that have nonlinear
weight functions based on look--up tables, that are specially smoothed in a
regularization called the diffusion. The idea of such a type of networks is
based on the hypothesis that the greater number of adaptive parameters per a
weight function might reduce the total number of the weight functions needed
to solve a given problem. Then, if the computational complexity of a
propagation through a single such a weight function would be kept low, then
the introduced neural networks might possibly be relatively fast.

A number of tests is performed, showing that the presented neural networks
may indeed perform better in some cases than the classic neural networks and
a number of other learning machines. 

\textbf{keywords:} feedforward neural networks, nonlinear regression,
nonlinear weight functions, generalization, diffusion
\end{abstract}

\section{INTRODUCTION}
Introducing adaptive nonlinear processing into the weight functions
of feedforward, densely connected neural network gives the network
the power of the order of \(N^{2}\) of adaptive nonlinear processing units, where \(N\)
is the number of nodes. For large \(N\), it can be a substantial difference
in comparison to the respective \(N\) adaptive nonlinear processing units in the classic
neural networks.

The feedforward neural networks presented in this paper have nonlinear weight
functions based on look--up tables -- a look--up table represents
nodes of a piecewise linear interpolation of the arguments of a
weight function. Thanks to this, only one or two nodes in the table
need to be read to propagate a signal.
This causes that only a chosen subset of the parameters is used during a
propagation of a given signal, what can make
the propagation time reasonable despite a very large number of adaptive parameters.
Another quality of the described neural networks is that the
subsequent parameters in the look--up table control propagation of subsequent
ranges in the domain of the weight function. Because of this, it can be
hypothesized that two subsequent adaptive parameters in the look--up table are
with some propability likely to control close points in the input space of the
neural network. An advantage from the property can be taken during additional
regularization of the weight function during the training process. Such a regularization
is very important in the case of the presented neural networks,
because it reduces the problems with the lack of smoothness of the look-up table
based adaptive functions, that may cause serious generalization problems
as reported by \cite{piazza1993lut}.

The regularization, called \emph{diffusion}, works as follows.
During the training process, the values of the propagated signals are traced to build
`visit' tables related to the frequency of falling of the look--up table
arguments into different ranges of values.
Each nonlinear weight function has such an accompanying visit table.
On the basis of the tables, a
regularization is performed, which `diffuses' values within the look--up
tables from the more
`visited' regions of the look--up tables to the less `visited' regions of
the look--up tables. This way, values of signals are extrapolated within
a weight function, and in effect it might be likely that the values
representing the generalized function are extrapolated by the discussed
neural networks as well.

Schematic examples of generalization in the cases of
a linear, spline--like and LUT weight functions are illustrated
in Fig.~\ref{fig:wftypes}.

        \begin{figure}[h!]
        \vskip 0.2in
        \begin{center}
        \begin{tabular}{cccc}
			\input wf_l-2.pstex_t &
			\input .pstex_t &
			\input V.pstex_t &
			\input wf-descr.pstex_t \\
			(a) & (b) & (c) & \\
		\end{tabular}
        \end{center}
        \caption{C}
        \label{fig:wftypes}
        \end{figure}

			{wf_s-2} 
			{wf_r-2} 
			{A schematic example of generalization using (a) linear, (b) smooth
			and (c) LUT
			weight function types. \(I\) are arguments and \(O\) are
			values of the weight functions, and crosses schematically denote points that
			minimize the training error function.}
The lack of smoothness of the weight function in (c) may clearly
decrease the generalization quality \cite{piazza1993lut}. Some adaptive
parameters are completely independent from the training samples in that case.
The weight function (b) is smooth and has a relatively small number
of adaptable parameters to improve generalization --
networks with similar, spline--like adaptive activation functions have been presented by
	Uncini, Capparelli and Piazza \cite{uncini1998},
	Vecci, Piazza and Uncini \cite{vecci1998learning}, and
	Guarnieri and Piazza \cite{guarnieri1999}.
Such a lower number of adaptive parameters, however, is exactly what we want to elude
in the proposed architecture.
An example of generalization using a weight function with a large number of adaptable parameters,
like in (c), without and with the diffusion of the function, is shown in
Fig.~\ref{fig:nn.mse.rm.1-1.circle.s3.l1.00.0.raster}(a)
and Fig.~\ref{fig:nn.mse.rm.1-1.circle.s3.l1.11.0.raster}(a), respectively.
It can be seen, comparing these figures and the respective generalizing functions,
that the diffusion may clearly improve the generalization.

The use of high number of adaptable parameters for each connection may raise the question
about bounds on the generalization performance of a learning machine
\cite{vapnik1995}. Yet, for first it should be noted, that the high number of the
parameters per connection might result in a lower total number of connections needed.
Secondly, even if the introduced neural networks might obviously have a very high
Vapnik Chervonenkis dimension, so the risk
bound would be very high, the VC dimension
takes into account only the maximum number of training points that can be
shattered by the learning machine. Thus
it clearly can be a very `loose' bound, in the sense that the other qualities
of the learning machine can make the actual risk much lower than
the bound. For example, the discussed diffusion can make the subsequent adaptable parameters
in the LUT weight function dependent on each other, yet in computing of
the VC dimension it is not taken into account at all. 

\section{NETWORKS WITH DIFFUSED WEIGHT FUNCTIONS}
\label{sec:net_structure}
\label{sec:neuron_model}
Let us separate the notion of a neuron into
a number of connections and a node.
We do so because there are two different types of
connections in the discussed neural networks.
The weight function is associated with a given connection,
and the combination and activation functions are associated with a node.
This way of describing a neuron is illustrated in Fig.~\ref{fig:neuron}

        \begin{figure}[h!]
        \smallskip
        \begin{center}
	\input neuron.pstex_t
        \end{center}
        \caption{}
        \label{fig:neuron}
        \end{figure}
VC
        {Structure of an example neuron.}

Let \(n\) denote the iteration number.
In a connection \(i\), the computation of the connection output value
\(O_{i}(n)\)
on the basis of the connection input value \(I_{i}(n)\) is done
by using the connection weight function. 

The combination function \(u^{c}_{k}(n)\) of a node \(k\) sums its arguments,
like in the classic neurons \cite{mcculloch43nervous}
\begin{equation}
\label{combination_function}
	u^{c}_{k}(n) = \sum_{i \in M}O_{i}(n),
\end{equation}
where \(M\) is a set of indexes of the input connections of the node \(k\).

The activation function \(u^{a}_{k}\) of a node \(k\) is sigmoid--like
\begin{equation}
\label{activation_function_tanh}
	y_{k}(n) = u^{a}_{k}(u^{c}_{k}(n)) = \tanh\Big(u^{c}_{k}(n)\Big),
\end{equation}
where \(y_{k}(n)\) is the output value of the node \(k\).
The activation function softly clamps the
combination function value, so that the value fits into the domains of
the LUT weight functions.

A linear connection \(i\) has a weight function of the form
\begin{equation}
\label{eq:scalar_weight_function}
	O_{i}(n) = w^{i}_{s}I_{i}(n),
\end{equation}
where \(w^{i}_{s}\) is the connection scalar weight.

A LUT connection \(i\) has the following weight function
\begin{equation}
\label{eq:combined_weight_function}
	O_{i}(n) = w^{i}_{l}(n)I_{i}(n) +
		r\Big(\mathbf{w}^{i}_{r}(n), I_{i}(n)\Big),
\end{equation}
where
\(w^{i}_{l}(n)I(n)\) is a component further called the linear one and
\(r(\mathbf{w}^{i}_{r}(n), I_{i}(n))\) is a component further called
the LUT one.
The coefficients \(w^{i}_{l}(n)\) and \(\mathbf{w}^{i}_{r}(n)\) are
the parameters of the connection weight function. The parameter
\(w^{i}_{l}(n)\) is a scalar and the parameter
\(\mathbf{w}^{i}_{r}(n)\) is the LUT of the weight function.
The function \(r(\mathbf{w}^{i}_{r}(n), I(n))\)
is      determined by a curve being
a linear interpolation of
several points \(p^{i}_{j}(n)(I^{j}, w^{i}_{r,j}(n))\),
where \(j = 0 \ldots r_{\mathrm{res}} - 1\) and
\(w^{i}_{r, j}(n)\) is a \(j\)th element of
\(\mathbf{w}^{i}_{r}(n)\). The first coordinate of the points on the curve
denotes arguments
and the second one values of the function.
The values \(I^{j}\) are equally distributed values as
follows
\begin{equation}
\label{eq:raster_function_argument_values}
	I^{j} = I_{\mathrm{min}} + \frac{j}
		{r_{\mathrm{res}} - 1}
		(I_{\mathrm{max}} - I_{\mathrm{min}}).
\end{equation}
Let us call
\(r_{\mathrm{res}}\) the LUT component resolution.
The coefficients \(I_{\mathrm{min}}\) and \(I_{\mathrm{max}}\) are
\(I_{i}(n)\)
minimum and maximum allowable values, respectively.
These values are equal to
the minimum and maximum values of the activation functions.
Let the function
\(r(\mathbf{w}^{i}_{r}(n), I(n))\) be computed using
the following piece--wise linear interpolation:
\begin{equation}
\label{eq:raster_function}
	\begin{array}{c}\displaystyle
		r\Big(\mathbf{w}^{i}_{r}(n), I(n)\Big) =
			\Big(\lfloor S(n) \rfloor + 1 - S(n)\Big)
				w^{i}_{r, \lfloor S(n) \rfloor}(n) +
			\Big(S(n) - \lfloor S(n) \rfloor\Big)
				w^{i}_{r, \lfloor S(n) \rfloor + 1}(n) \\[12pt]
		\displaystyle
		S(n) = \frac{I(n) - I_{\mathrm{min}}}
			{I_{\mathrm{max}} - I_{\mathrm{min}}}
			\left(r_{\mathrm{res}} - 1\right)
		. \\
	\end{array}
\end{equation}

Alternatively, of course,
another interpolation type, for example cubic spline interpolation
\cite{greville1969spline, deboor1978splines}, could be used.

The memory overhead of the presented networks grows
linearly with \(r_{\mathrm{res}}\) -- each nonlinear
weight function needs only one additional table for
the diffusion process discussed in
Sec.~\ref{sec:lut_regularization}, and the additional table
is of the size of \(r_{\mathrm{res}}\).
Thus, even with networks having thousands of
connections, and \(r_{\mathrm{res}}\) being of the order of hundreds, the
overhead can be low on modern computers, because such computers
often feature hundreds of megabytes of memory.

Because the piecewise linear interpolation requires only
one or two adaptive parameters, the ratio of the number of the used parameters
within a single propagation to the number of all of the parameters is less that or equal to
\(2/r_{\mathrm{res}}\). Let us call the quality of using only some chosen adaptive
parameters the selective parameters property. The property can make propagation
very fast, while still retaining a large number of adaptive parameters available to
the training process. The relatively fast propagation in the presented networks will be
demonstrated in tests.
If, instead of the piecewise interpolation, a function like
a single polynomial would be used for a weight functions, then the property of
selective parameters would obviously not apply, because each adaptive parameter
would be needed to find a value of the function.

The linear component \(w^{i}_{l}(n)I_{i}(n)\) has the role of generalizing
linear patterns. It has been found in tests that a neural network with
both linear and the LUT components may in some cases perform
substantially better than a network with only the LUT components.

The introduced networks are fully connected multilayer feedforward neural networks
\cite{bishop1995neural, hertz1991neural}.
All linear connections that would be used in a basic
layered feedforward neural network, except of these from bias elements, are
replaced with LUT connections in the presented networks.
Bias elements have a constant output, and
therefore there is no need for a LUT connection.
A sample such a NN is illustrated in Fig.~\ref{fig:lfnn}.

        \begin{figure}[h!]
        \smallskip
        \begin{center}
	\input lfnn.pstex_t
        \end{center}
        \caption{}
        \label{fig:lfnn}
        \end{figure}
VC
    {An example of a fully connected multilayer feedforward NN
     with nonlinear weight functions, \(L_{i}\) denotes the \(i\)th layer.}

\section{THE LEARNING ALGORITHM}
The presented NNs have their distinct learning algorithm, consisting of a
modified error backpropagation\cite{rumelhart1996parallel, bishop1995neural, hertz1991neural}
and a regularization of the weight functions.

A schematic diagram of a single iteration of the learning algorithm is presented in
Fig.~\ref{fig:nlw-learning}.

        \begin{figure}[h!]
        \smallskip
        \begin{center}
	\input nlw-learning.pstex_t
        \end{center}
        \caption{}
        \label{fig:nlw-learning}
        \end{figure}
VC
    {Diagram of a single iteration of the learning algorithm.}
In the beginning of an iteration, attributes of the training samples
are propagated through the network.
The propagation needs derivatives of the weight functions. In the case of the
LUT weight function, approximation of the derivative is computed instead.
During the propagation, weight functions are adapted and the visit functions are updated.
Then, regularization is performed. 
Within the regularization, weight decay is performed on both the linear and nonlinear weight
functions. Also, the nonlinear weight functions and the visit functions are diffused,
according to the values in the visit functions.

Note that the order of the mentioned operations is not critical --
the training process may have many iterations, and so the progressive changes
within a single training iteration are usually relatively low. In the detailed
description of the learning algorithm later in this section, for mathematical
completeness, the blocks presented in Fig.~\ref{fig:nlw-learning} are
tied in a specific order, but the order is practically unimportant.

\subsection{TRAINING}
\label{sec:net_training}
On--line training with error backpropagation
\cite{rumelhart1996parallel, bishop1995neural, hertz1991neural} is used.

Let a training set
be given. Each sample \(k\) in the set has \(i\) argument attributes
and \(j\) value attributes
\(\left\{x^{k}_{0}, x^{k}_{1}, \ldots x^{k}_{i - 1},
d^{k}_{0}, d^{k}_{1}, \ldots d^{k}_{j - 1}\right\}\)
and we want the network to generalize the relation between the
attributes with a function mapping the argument attributes to
the value attributes.
Let \(e_{i}\) be an error function derivative backpropagated
to the connection i, and \(\mu\) be the learning step.

	To keep
	the descriptions of the training algorithm and the regularization
	algorithm separate, here
	the regularization functions \(R_{d}(w)\)
	and \(R_{s}(w, \Delta w)\) will only be briefly mentioned,
	and described in detail in Sec.~\ref{sec:regularization}.
    The function \(R_{d}(w)\) is a simple weight decay
    \cite{krogh1992simple} -- its value is its
 	argument multiplied by a value in the range \(\left<0, 1\right>\).
    The function \(R_{s}(w, \Delta w)\)
    is a kind of an indirect weight decay, that does not have some
    drawbacks of the regular weight decay, but let us now for
    simplicity assume that
    \begin{equation}
    	R_{s}(w, \Delta w) = \Delta w
    	.
    \end{equation}
 
 	The regularization of the LUT component
 	can be much more computationally complex that the
 	regularization of the linear connection and of the linear component,
 	because of the number of adaptive parameters of the LUT component.
 	Because of this, while the regularization of the linear connection and
 	of the linear component is performed
 	in every training iteration, the regularization of the LUT component
 	is performed only in some training iterations, in which
 	the following is true:
\begin{equation}
	\label{eq:lut-regularization_exclusion}
	\zeta > \zeta_{i}(n),
	\label{eq:random-regularization-exclusion}
\end{equation}
where
\begin{equation}
	\zeta_{i}(n) = \mathrm{rand}(0.0, 1.0),
\end{equation}
and the function \(\mathrm{rand}(0, 1)\) is a uniform random number
generator which returns a random value within \(\left<0, 1\right)\).
The more the \(\zeta\) coefficient is lower than \(1\),
the less is the mean computation complexity per training iteration,
but the quality of the regularization may be worse. The LUT
regularization can accordingly be `stronger' to counterbalance
its exclusion in some iterations.
The random exclusion (\ref{eq:lut-regularization_exclusion}),
instead of a regular one, is used to rule out possible
resonance with the training samples.
Condition (\ref{eq:lut-regularization_exclusion}) will
be repeated in some equations in the later sections.

\subsubsection{ADAPTING WEIGHTS OF THE LINEAR CONNECTIONS}
The weights are randomly initialized before training.
with values in the range \(\left<-0.5, 0.5\right>\).

Let there be a linear connection \(i\) with
the input value \(I_{i}(n)\).
The weight of the linear connection is adapted as in the classic
propagation, with the weight decay applied:
\begin{equation}
	\label{eq:linear_weight_update}
	w^{i}_{s}(n + 1) = R_{d}\Big(w^{i}_{s}(n) +
		\Delta w^{i}_{s}(n)\Big),
\end{equation}
where
\begin{equation}
\label{eq:linear_weight_difference}
	\Delta w^{i}_{s}(n) = - R_{s}\bigg(
		w^{i}_{s}(n),
		\mu e_{i}(n)I_{i}(n)\bigg).
\end{equation}

\subsubsection{ADAPTING WEIGHTS OF THE LUT CONNECTIONS}
\label{sec:updating_raster_weight}
A LUT component is randomly initialized before the training of the neural network
with small values in the range \(\left<-0.5, 0.5\right>\) and a constant derivative.

Let there be a LUT connection \(i\). Let the linear component weight 
\(w^{i}_{l}(n)\) be adapted analogously to the weight in the linear connection:
\begin{equation}
	\label{eq:linear_component_update}
	w^{i}_{l}(n + 1) = R_{d}\Big(w^{i}_{l}(n) +
		\nu \Delta w^{i}_{l}(n)\Big),
\end{equation}
where
\begin{equation}
\label{eq:linear_component_difference}
	\Delta w^{i}_{l}(n) = - R_{s}\bigg(
		w^{i}_{l}(n),
		\mu e_{i}(n)I_{i}(n)\bigg).
\end{equation}
The coefficient \(\nu\)
specifies a relation between the adaptation speed of the linear component and
the adaptation speed of the LUT component. Increasing the value may cause the 
linear patterns to have a greater impact on the generalizing function.

Let the LUT component weight
\(\mathbf{w}^{i}_{r}(n)\)
be adapted also in a similar way of that of the linear connection:
\begin{equation}
\label{eq:weight_raster_component_update}
\begin{array}{c}
	r\Big(\mathbf{w}^{i**}_{r}(n + 1), I_{i}(n)\Big) =
	r\Big(\mathbf{w}^{i}_{r}(n), I_{i}(n)\Big) + \Delta w^{i}_{r}(n)
	\\
	
	r\Big(\mathbf{w}^{i*}_{r}(n + 1), I_{i}(n)\Big) =
	\left\{
	\begin{array}{ll}
		R_{d}\bigg(
		r\Big(\mathbf{w}^{i**}_{r}(n), I_{i}(n)\Big)
		\bigg) &
		\textrm{if \(\zeta > \zeta_{i}\)} \\
		r\Big(\mathbf{w}^{i**}_{r}(n), I_{i}(n)\Big)
		&
		\textrm{if \(\zeta \le \zeta_{i}\)} \\
	\end{array}
	\right.
	\\
\end{array}
	,
\end{equation}
where
\begin{equation}
\label{eq:raster_weight_difference}
	\Delta w^{i}_{r}(n) = - R_{s}\bigg(
		r\Big(\mathbf{w}^{i}_{r}(n), I_{i}(n)\Big),
		\mu e_{i}(n)\bigg)
	.
\end{equation}
Because the value of the LUT component is not
a product of \(I_{i}\) and of
\(r(\mathbf{w}^{i}_{r}(n), I_{i}(n))\),
but is the value \(r(\mathbf{w}^{i}_{r}(n), I_{i}(n))\) itself,
the term \(e_{i}(n)\) is used in
(\ref{eq:raster_weight_difference}) instead of the term
\(e_{i}(n)I_{i}(n)\) as in (\ref{eq:linear_weight_difference}).
The symbol \(^{*}\) in
(\ref{eq:raster_weight_difference}) and
later in this section denotes a value before the diffusion of the
LUT component. The diffusion which is described later
in Sec.~\ref{sec:lut_regularization}.

The equation (\ref{eq:weight_raster_component_update}) changes
a value of the LUT component, but it does not decompose the
change on the individual values in the LUT.
Let the following conditions be given on the adaptation of the LUT.
If \(I_{i}(n)\) is equal to a \(I^{j}\) coordinate
of an approximated point \(p^{i}_{j}(I^{j}, w^{i}_{r,j}(n))\), then
only that point value \(w^{i}_{r,j}(n)\) is changed. To fulfill
(\ref{eq:weight_raster_component_update}),
\begin{equation}
	w^{i**}_{r,j}(n + 1) =  w^{i}_{r,j}(n) +
		\Delta w^{i}_{r}(n).
\end{equation}
Otherwise, 
\(\exists j \in \left<0, r_{\mathrm{res}} - 2\right>\),
\(I_{i}(n) \in (I^{j}, I^{j + 1})\). 
Then, values
\(w^{i}_{r,j}\) and \(w^{i}_{r,j + 1}\) are changed, using
the amounts \(\Delta w^{r}_{L}(n)\) and \(\Delta w^{r}_{H}(n)\),
respectively,
\begin{equation}
	\begin{array}{c}
		w^{i**}_{r,j}(n + 1) =
			w^{i}_{r,j}(n) + \Delta w^{r}_{L}(n) \\[8pt]
		w^{i**}_{r,j + 1}(n + 1) =
			w^{i}_{r,j + 1}(n) + \Delta w^{r}_{H}(n) \\
	\end{array}
	,
\end{equation}
such that
\begin{equation}
	\label{eq:weight_change_proportion}
	\frac{\Delta w^{r}_{H}(n)}{\Delta w^{r}_{L}(n)} =
		\frac{I_{i}(n) - I^{j}(n)}
			{I^{j + 1}(n) - I_{i}(n)}.
\end{equation}
Therefore, one of the modified points, denoted \(k\),
whose \(I^{k}\) value is possibly nearer to
\(I_{i}(n)\), has its \(w^{i}_{r,k}\) value changed by a greater
amount.
To fulfill (\ref{eq:weight_raster_component_update}) and
(\ref{eq:weight_change_proportion}),
\(\Delta w^{r}_{L}(n)\) and \(\Delta w^{r}_{H}(n)\) have
the following form
\begin{equation}
	\begin{array}{c}\displaystyle
		\Delta w^{r}_{L}(n) = \Delta w^{i}_{r}(n)\frac{r_{L}}
			{2r_{L}^{2} - 2r_{L} + 1} \quad
				r_{L} = I^{i}_{j + 1}(n) - I_{i}(n) \\[14pt]
		\displaystyle
		\Delta w^{r}_{H}(n) = \Delta w^{i}_{r}(n)\frac{r_{H}}
			{2r_{H}^{2} - 2r_{H} + 1} \quad
				r_{H} = I_{i}(n) - I^{i}_{j}(n) \\
	\end{array}
	.
\end{equation}

As can be seen, the quality of selective parameters apply also to
the adapting of the parameters -- only one or two parameters are
modified during training within a single iterations, independently
of the value of \(r_{\mathrm{res}}\).

\subsubsection{APPROXIMATED DERIVATIVE OF LUT WEIGHT FUNCTION}
\label{sec:approximated_derivative}
For error backpropagation to work, a derivative
of a weight function in respect to a connection input value is
needed. In the case of a LUT connection,
an approximation of the derivative will be used instead.
Let it be described as follows
\begin{equation}
	\frac{\partial v\Big(\mathbf{w}_{i}(n),
		I_{i}(n)\Big)}
		{\partial I_{i}(n)} =
	c^{i}_{l}(n) + c^{i}_{r}(n),
\end{equation}
where \(c^{i}_{l}(n)\) and \(c^{i}_{r}(n)\) are a derivative
of the linear component and approximated derivative of the LUT
component, respectively. Of course,
\begin{equation}
	c^{i}_{l}(n) = w^{i}_{l}(n).
\end{equation}
In the case of \(c^{i}_{r}(n)\), a derivative approximation
evaluated as the difference of neighboring LUT values is not used,
because it could be too sensitive to individual
weight changes and
could cause numerical instability \cite{guarnieri1999multilayer}.
Instead, the approximated derivative of a LUT component
\(r\) of a connection \(i\) is given by
\begin{equation}
\label{eq:approximated_derivative}
	\begin{array}{c}\displaystyle
	c^{i}_{r}(n) = \parallel A\parallel^{-1}\sum_{a \in A}
		\frac{r\Big(\mathbf{w}^{i}_{r}, c(I_{i}(n) + a)\Big) -
			r\Big(\mathbf{w}^{i}_{r}, c(I_{i}(n) - a)\Big)}
			{c(I_{i}(n) + a) - c(I_{i}(n) - a)}
	\\[19pt]
	\displaystyle
	A = \{a_{l}, a_{m}a_{l}, a^{2}_{m}a_{l}, \ldots a_{f}\}
	\quad a_{f} \le a_{h} \quad a_{m}a_{f} > a_{h}
	\\[11pt]
	\displaystyle
	c(q) = \left\{\begin{array}{ll}
		I_{\mathrm{min}} & \textrm{if
			\(q < I_{\mathrm{min}}\)} \\
		q & \textrm{if
			\(I_{\mathrm{min}} \le q \le I_{\mathrm{max}}\)} \\
		I_{\mathrm{max}} & \textrm{if
			\(q > I_{\mathrm{max}}\)} \\
	\end{array}\right.
	\end{array}
	.
\end{equation}
Therefore, the approximated derivative
is the mean of several differential
ratios of the LUT component.
The coefficient \(a_{l}\) is the
minimum value of \(a\), the coefficient \(a_{h}\) is
the maximum value of \(a\) that possibly exists,
and \(a_{m}\) determines the number of the ratios.

\subsection{REGULARIZATION}
\label{sec:regularization}
Two types of regularization are used in the proposed neural
networks -- of absolute values of weight functions and
of the diffusion of the LUT components.

\subsubsection{REGULARIZATION OF ABSOLUTE VALUES OF WEIGHT FUNCTIONS}
\label{sec:regularization-absolute-values}
This type of regularization is a kind of weight decay, that tries to
prevent absolute values of the weight functions
from getting too large. 
Weight decay can improve generalization \cite{krogh1992simple}.
Regularization is used for
the adaptable parameter \(w^{i}_{s}\) of a linear connection,
the linear component adaptable parameter \(w^{i}_{l}\) and the LUT component
adaptable parameters \(\mathbf{w}^{i}_{r}\).
To regularize weights, the functions \(R_{s}(w, \Delta w)\) and
\(R_{d}(w)\) are used.
The functions were already used in the equations in
Sec.~\ref{sec:net_training}.
In this section, the functions will be described in more detail.

Let the function \(R_{d}(w)\) be as follows
\begin{equation}
\label{eq:weight_diminishment}
 R_{d}(w) = (1 - R^{s}_{b})w
 .
\end{equation}
As can be seen, it is a simple weight decay,
whose strength is determined by the coefficient \(R^{s}_{b}\).

A weight decay like in (\ref{eq:weight_diminishment}) may have the
disadvantage of preventing the training process of converging
exactly into a local minimum -- the decay always `pushes' the weights toward
zero. Yet this type of regularization can still be very important
\cite{krogh1992simple}. In the presented algorithm, the following
solution is proposed. The weight decay (\ref{eq:weight_diminishment}),
should it be required to be too strong,
is partially substituted by another type of weight decay, that
operates not directly on the values of weights, but instead
on the gains of the weights.
Let the another type of weight decay be represented by the
function \(R_{s}(w, \Delta w)\), that has the following equation:
\begin{equation}
	R_{s}(w, \Delta w) = \left\{
		\begin{array}{ll}\displaystyle
			\frac{\exp(R^{s}_{a} w \Delta w) - 1}
				{R^{s}_{a} w} &
				\textrm{if \(w \ne 0 \land R^{s}_{a} \ne 0\)} \\[14pt]
			\displaystyle
			\Delta w &
				\textrm{if \(w = 0 \lor R^{s}_{a} = 0\)} \\
		\end{array}
	\right.
\end{equation}
where \(R^{s}_{a}\) determines the level of the regularization.
As can be seen, the more \(w\) is positive, the more the ratio
\(\frac{R_{s}(w, \Delta w)}{\Delta w}\) decreases as
\(\Delta w\) increases,
and conversely, the more \(w\) is negative, the more the ratio
\(\frac{R_{s}(w, \Delta w)}{\Delta w}\) increases as
\(\Delta w\) increases.
That type of regularization may both slow down increasing of absolute
values of the weight functions and speed up decreasing the
absolute values.

\subsubsection{DIFFUSION OF THE LUT COMPONENT}
\label{sec:lut_regularization}

A given training sample can, by its very presence, make
the point that it represents in the input space of the regressor,
and the surroundings of the point, more statistically significant
or defined. The `spreading' of the significance over the surroundings is used in
methods like for example the nearest neighbor or cubic spline interpolation
\cite{greville1969spline, deboor1978splines}.
Assuming that the weight functions are not very
`jagged', it can be hypothesized that two subsequent adaptive parameters
in the look--up table are likely to control close points in the input space of
the neural network.
The idea behind diffusion is based just on that. A sample has a `significance'
and while its attributes are propagated through the network, the
`significance' is marked in respective regions of the weight functions by
the means of the accompanying visit function.
In the diffusion process, values in the more `significant' regions of the
weight functions are heuristically `diffused' to the less `significant'
regions of the weight function, thus heuristically performing an interpolation
by the `spreading' of the significance of samples like in the mentioned nearest neighbor or
cubic spline interpolations.

The algorithm of diffusion was constructed so as to have low memory requirements
and low time complexity. For each LUT component it needs only one additional
visit table -- of the size of LUT of the component, and the time complexity is roughly
linear to the resolution of the LUT. On the other hand, the algorithm is
very far from Fick's diffusion equation, and the `significance'
estimation is heuristic. The algorithm, however,
keeps the time of a single iteration relatively short. This may be important
if there are many training samples, and in effect many iterations
to propagate the attributes of the samples are required.

In the process of diffusion, roughly speaking, the weighted density of occurrence of
LUT component arguments, at different regions of the LUT component domain,
is computed. In computing the density, there is a higher importance given 
to the more recent iterations.
There are two reasons for giving the more recent iterations a higher
importance. First, because of the adaptation of weight functions
during training, the way of propagation of signals may gradually change,
and we want the weight functions to fit to the more `current' way of
propagation of the signals.
Secondly, because we use an on--line learning method, there may
be possible trends in the training data.

Values related to the densities are stored in the visit table.
For each single value in the LUT weight function, there is a single
corresponding value in the visit table. Values within the
weight function LUT having the relatively higher
corresponding values in the visit table are `diffused' to these neighboring
ones that have the relatively lower corresponding values in the visit table.
Let the mean
of two such neighboring values of the weight function LUT
before diffusion be \(m\). After diffusion, the value
that had higher
corresponding value in a visit table moves less towards \(m\) than the
other value.
The diffusion
`spans' incrementally the LUT component function in regions relatively
less frequently modified or not modified at all, where the 
`spanning' regions are these relatively more frequently modified. 
There is no binary division only into extreme `spanning' and `spanned'
regions, of course, as the visit functions are multivalued.

The diffusion is also applied to the visit table.
This is because if
a value `diffuses' to a neighboring one, an `importance' of the value
`diffuses' also.

During the diffusion process, the LUT functions are also
`smoothed' by decreasing the absolute differences between
neighboring values in the LUTs, to reduce the problems
caused by the lack of smoothness as reported in
\cite{piazza1993lut}.

Let there be two subsequent values \(w^{i*}_{r,j}(n + 1)\) and
\(w^{i*}_{r,j + 1}(n + 1)\) of a LUT component
\(r(\mathbf{w}^{i*}_{r}(n + 1),I_{i}(n))\), as described in
Sec.~\ref{sec:net_training}. We want to smooth
\(r(\mathbf{w}^{i*}_{r}(n + 1),I_{i}(n))\) by making the absolute
difference \(|w^{i}_{r,j + 1}(n + 1) - w^{i}_{r,j}(n + 1)|\) smaller
than \(|w^{i*}_{r,j + 1}(n + 1) - w^{i*}_{r,j}(n + 1)|\).
We also possibly want to
`diffuse' each \(w^{i*}_{r,j}(n + 1)\) value to
\(w^{i}_{r,j - 1}(n + 1)\) and
\(w^{i}_{r,j + 1}(n + 1)\), depending on visit table values.
Let a LUT element
\(w^{i}_{r,j}(n)\), \(j = 0 \ldots r_{\mathrm{res}} - 1\),
have its associated visit table
element \(V^{i}_{j}(n)\).
Let \(V^{i*}_{j}(n + 1)\) be the visit function values before
LUT component regularization.
The following equation fulfills the discussed assumptions:
\begin{equation}
\label{eq:raster_weight_regularizing}
\begin{array}{c}
	\begin{array}{c}\displaystyle
		w^{i,L}_{r,j}(n + 1) = m_{j} -
			\frac{\tanh(R^{r}_{a}d_{j})}
			{2R^{r}_{a}(1 + R^{r}_{b}p_{j})} \quad
			0 \le j \le r_{\mathrm{res}} - 2 \\[14pt]
		\displaystyle
		w^{i,H}_{r,j + 1}(n + 1) = m_{j} +
			\frac{\tanh(R^{r}_{a}d_{j})}
			{2R^{r}_{a}(1 + R^{r}_{b}p_{j}^{-1})} \quad
			0 \le j \le r_{\mathrm{res}} - 2 \\[20pt]
		\displaystyle
		d_{j} = w^{i*}_{r,j + 1}(n + 1) - w^{i*}_{r,j}(n + 1) \\[10pt]
		\displaystyle
		m_{j} = \Big(w^{i*}_{r,j}(n + 1) + w^{i*}_{r,j + 1}(n + 1)\Big)/2
		\\[10pt]
		\displaystyle
		p_{j} = \frac{V^{i*}_{j + 1}(n + 1)}{V^{i*}_{j}(n + 1)} \\[18pt]
		\displaystyle
		w^{\prime i}_{r,0}(n + 1) = w^{i,L}_{r,0}(n + 1) \\[14pt]
		\displaystyle
		w^{\prime i}_{r,j}(n + 1) = \Big(w^{i,L}_{r,j}(n + 1) +
			w^{i,H}_{r,j}(n + 1)\Big)/2 \quad
			1 \le j \le r_{\mathrm{res}} - 2 \\[14pt]
		\displaystyle
		w^{\prime i}_{r,r_{\mathrm{res}} - 1}(n + 1) =
			w^{i,H}_{r,r_{\mathrm{res}} - 1}(n + 1) \\[14pt]
	\end{array}
	\\
	\forall_{i = 0, 1, \ldots r_{\mathrm{res}} - 1} \quad
	w^{i}_{r,j}(n + 1) = \left\{
	\begin{array}{ll}
		w^{\prime i}_{r,j}(n + 1)
		& \textrm{if \(\zeta > \zeta_{i}(n)\)} \\
		w^{i*}_{r,j}(n + 1)
		& \textrm{if \(\zeta \le \zeta_{i}(n)\)} \\
	\end{array}
	\right.
	. \\
\end{array}
\end{equation}
The coefficients
\(R^{r}_{a}\) and \(R^{r}_{b}\) determine a smoothing level
and a diffusion speed, respectively.
To increase numerical precision, the term \(p_{j}^{-1}\) is
computed directly from the \(V^{i*}_{j}\) values, which is important
because these values may get extremely low.
The computation of the two values \(w^{i,L}_{r,j}(n + 1)\) and
\(w^{i,R}_{r,j}(n + 1)\), and then computing of their mean, with the
exception of special cases at the two LUT values
having indexes \(0\) and \(r_{\mathrm{res}} - 1\), is performed
to maintain symmetry of the regularization of the LUT.

\subsubsection{COMPUTING THE VISIT TABLE}
Let us finally discuss computing the values in the visit table.
Let the values \(V^{i*}_{j}(n)\), that is the values
of the visit table before a
possible diffusion, have an equation as follows
\begin{equation}
	\label{eq:computing_visit_table}
	\begin{array}{c}\displaystyle
		V^{i*}_{j}(0) = V_{p} 
			\\[7pt]
		\displaystyle
		V^{i*}_{j}(n + 1) = \left\{\begin{array}{ll}\displaystyle
			l\Big((1 - R^{r}_{c})V^{i}_{j}(n)\Big) & \textrm{
			if \(\Big(j \ne \lfloor S_{i}(n + 1) \rfloor\Big)
				\land\)} \\[5pt]
			\displaystyle
			& \textrm{\(\quad \land
				\Big(j \ne \lceil S_{i}(n + 1) \rceil
				\Big)\)} \\[14pt]
			\displaystyle
			\begin{array}{l}
			\displaystyle
			l\Big((1 - R^{r}_{c})V^{i}_{j}(n)\Big)\bigg(1 + \\
			\displaystyle
				\qquad + R^{r}_{c}\Big(\lfloor S_{i}(n + 1)
					\rfloor + 1 - \\
			\displaystyle
				\qquad - S_{i}(n + 1)\Big)
				\Big(1 - V^{i}_{j}(n)\Big)\bigg) \\
			\end{array}
			& \textrm{
			if \(j = \lfloor S_{i}(n + 1) \rfloor\)} \\[14pt]
			\displaystyle
			\begin{array}{l}
			\displaystyle
			l\Big((1 - R^{r}_{c})V^{i}_{j}(n)\Big)\bigg(1 + \\
			\displaystyle
				\qquad + R^{r}_{c}\Big(S_{i}(n + 1) - \\
			\displaystyle
				\qquad - \lfloor S_{i}(n + 1) \rfloor\Big)
				\Big(1 - V^{i}_{j}(n)\Big)\bigg) \\
			\end{array}
			& \textrm{
			if \(j = \lceil S_{i}(n + 1) \rceil\)} \\
		\end{array}\right. \\
		\\[-4pt]
		l(x) = \max\left(
			x, V_{\mathrm{min}}\right)\\[7pt]
		\displaystyle
		S_{i}(n + 1) = \frac{I_{i}(n) - I_{\mathrm{min}}}
			{I_{\mathrm{max}} - I_{\mathrm{min}}}
			\left(r_{\mathrm{res}} - 1\right) \\[14pt]
		j =  0 \ldots r_{\mathrm{res}} - 1 \quad.\\
	\end{array}
\end{equation}
where
\(S_{i}(n + 1)\) scales \(I_{i}(n)\) like in
(\ref{eq:raster_function}). The coefficient \(V_{p}\) is the initial
value.
The coefficient \(V_{\mathrm{min}}\) has a very small positive value
and is used because of the limited precision of the representation
of real numbers used in computers.
The coefficient \(R^{r}_{c}\) determines
how large is the loss of importance of the less recent iterations
in computing of the values of the visit table.

The values obtained from (\ref{eq:computing_visit_table}) are used
in diffusing the weight function LUT, as was shown
in (\ref{eq:raster_weight_regularizing}), yet the visit table
is also diffused, because of the reasons already discussed
in Sec.~\ref{sec:lut_regularization}.
The visit table is diffused like the weight function LUT, but
without smoothing -- it was decided to omit the
smoothing here, because, in contrast to the weight function,
the visit table does not directly affect the propagation of signals.
Thus, the following formula is used for diffusing the visit table:
\begin{equation}
\label{eq:visit_table_regularizing}
\begin{array}{c}
	\begin{array}{c}\displaystyle
		V^{i,L}_{j}(n + 1) = m_{j} -
			\frac{d_{j}}
			{2(1 + R^{r}_{b}p_{j})} \quad
			0 \le j \le r_{\mathrm{res}} - 2 \\[14pt]
		\displaystyle
		V^{i,H}_{j + 1}(n + 1) = m_{j} +
			\frac{d_{j}}
			{2(1 + R^{r}_{b}p_{j}^{-1})} \quad
			0 \le j \le r_{\mathrm{res}} - 2 \\[20pt]
		\displaystyle
		d_{j} = V^{i*}_{j + 1}(n + 1) - V^{i*}_{j}(n + 1) \\[10pt]
		\displaystyle
		m_{j} = \Big(V^{i*}_{j}(n + 1) + V^{i*}_{j + 1}(n + 1)\Big)/2
		\\[10pt]
		\displaystyle
		p_{j} = \frac{V^{i*}_{j + 1}(n + 1)}{V^{i*}_{j}(n + 1)} \\[18pt]
		\displaystyle
		V^{\prime i}_{0}(n + 1) = V^{i,L}_{0}(n + 1) \\[14pt]
		\displaystyle
		V^{\prime i}_{j}(n + 1) = \Big(V^{i,L}_{j}(n + 1) +
			V^{i,H}_{j}(n + 1)\Big)/2 \quad
			1 \le j \le r_{\mathrm{res}} - 2 \\[14pt]
		\displaystyle
		V^{\prime i}_{r_{\mathrm{res}} - 1}(n + 1) =
			V^{i,H}_{r_{\mathrm{res}} - 1}(n + 1) \\[14pt]
	\end{array}
	\\
	\forall_{i = 0, 1, \ldots r_{\mathrm{res}} - 1} \quad
	V^{i}_{r,j}(n + 1) = \left\{
	\begin{array}{ll}
		V^{\prime i}_{j}(n + 1)
		& \textrm{if \(\zeta > \zeta_{i}(n)\)} \\
		V^{i*}_{j}(n + 1)
		& \textrm{if \(\zeta \le \zeta_{i}(n)\)} \\
	\end{array}
	\right.
	. \\
\end{array}
\end{equation}

\section{TESTS}
\label{sec:applications}
In this section, the presented networks will be tested and compared to some
other neural and non neural learning machines.

Unless otherwise stated, the following coefficients, selected
in a number of preliminary trials, are used in
the tests in this section:
\(\mu = 0.02\),
\(\nu = 2.5\),
\(r_{\mathrm{res}} = 64\),
\(I_{\mathrm{min}} = -1\),
\(I_{\mathrm{max}} = 1\),
\(a_{l} = 0.15\),
\(a_{h} = 0.35\),
\(a_{m} = 1.1\),
\(\zeta = 0.05\),
\(R^{r}_{a} = 1 \cdot 10^{-4}\),
\(R^{r}_{b} = 1 \cdot 10^{-4}\),
\(V_{p} = 0.1\),
\(V_{\mathrm{min}} = 1 \cdot 10^{-16}\),
\(R^{r}_{c} = 0.001\).
For the classic neural networks with linear weight functions only,
to make their weight decay like the classic one described in \cite{krogh1992simple},
\(R_{s}(w, \Delta w)\) is linear because
\(R^{s}_{a} = 0\), and the
weight decay coefficient \(R^{s}_{b}\) is equal to \(2 \cdot 10^{-7}\).
For the networks with diffused weight functions
\(R_{s}(w, \Delta w)\) is nonlinear and regularizes weight change
at \(R^{s}_{a} = 1\),
but the weight decay is weaker instead -- the coefficient
\(R^{s}_{b}\) is equal to \(1 \cdot 10^{-9}\).
To show the regularization capabilities of the presented networks,
only the \(\mu\), \(r_{\textrm{res}}\) and \(R^{r}_{b}\) coefficients will
actually be fitted to different
generalized data sets, except for some special cases,
and the rest of the coefficients will be constant.

The `LW' prefix is used to represent the classic
feedforward networks with linear weight functions only, whereas
the `NLW' prefix corresponds to the introduced networks.
Following this prefix is the number of nodes in the subsequent layers,
after the input layer.
Thus, a classic neural network named LW 2--4--4--1
would have 2 nodes in the input layer,
followed by two layers
with 4 nodes in each, and finally a single node in
the output layer.

\subsection{TIME COMPLEXITY}
The speed of a signal propagation through a connection and the time of
modifying a connection weight can be substantially different for
connections with the linear and the LUT weights. In this section, some time
measurements are provided for estimation and comparison of the learning and
propagation performance of both the classic and the introduced networks. The
time results are for a particular implementation and should be interpreted
with care, of course.

A number of architectures was tested, with different number of inputs,
outputs, layers and nodes within a layer.
The iteration time to the number of connections ratio was
generalized by linear functions of the form \(a + bn\), as shown in Table
\ref{tab:time-complexity},
where \(n\) is the number of connections, the time is in milliseconds
and the +/- values indicate the lower and upper parallel bounding
lines of the measured times, respectively.
\begin{table}
\begin{center}
\begin{tabular}{l|cc}
\(a + bn\)				&	a							&		b	\\[6pt]
\hline
\hline\\[-3pt]
LW training				&	\(-0.81_{-1.59}^{+2.56}\)	&	\(0.013\)	\\[3pt]
\hline\\[-4pt]
LW propagation only		&	\(-0.28_{-0.58}^{+1.42}\)	&	\(0.004\)	\\[3pt]
\hline\\[-4pt]
NLW training			&	\(-1.84_{-2.07}^{+4.30}\)	&	\(0.056\)	\\[3pt]
\hline\\[-4pt]
NLW propagation only	&	\(-0.39_{-0.84}^{+1.20}\)	&	\(0.010\)	\\[3pt]
\hline\\[-4pt]
\end{tabular}
\caption{Time complexity for a single iteration generalized using linear regression.}
\label{tab:time-complexity}
\end{center}
\end{table}
The difference
in the iteration only times is only of about 2.5 times, what results
from using a fast piecewise linear interpolation, resulting in the
selective parameters property, in the nonlinear weight
functions.

The fitted function for learning times against \(r_{\mathrm{res}}\) is as follows
\[
	\textrm{nrl}(r_{\mathrm{res}}) = 38.1_{-66.6}^{+310} + 0.113r_{\mathrm{res}}
\]
For \(r_{\mathrm{res}}\) increase of 16 times, from 16 to 256,
there is a
time increase per iteration of about \(1.7\). It is
because of the selective parameters property of the weight functions,
and because \(\zeta = 0.05 \ll 1\) so \(r_{\mathrm{res}}\) value is critical for time
performance in only about 5\% of all training iterations.

\newcommand{\rasterdiagramsoutputweights}[4] {
	\psfigureabc{nn.mse.rm.1-1.#1.#3.00.0.images}
		{nn.mse.rm.1-1.#1.#3.00.0.image.output.1000}
		{nn.mse.rm.1-1.#1.#3.00.0.image.output.10000}
		{nn.mse.rm.1-1.#1.#3.00.0.image.output.100000}
		{1.0in}
		{Images of the generalizing function after
		(a) \(1000\), (b) \(10000\) and (c) \(1\cdot 10^{5}\) iterations,
		respectively, of a
		1--1 NN for an image `#2', \(R^{r}_{a} = 0\),
		\(R^{r}_{b} = 0\).
		}
	\psfigureabcdef{nn.mse.rm.1-1.#1.#3.00.0.raster}
		{nn.mse.rm.1-1.#1.#3.00.0.1.raster.diagram}
		{nn.mse.rm.1-1.#1.#3.00.0.1.raster.wmi.diagram}
		{nn.mse.rm.1-1.#1.#3.00.0.2.raster.diagram}
		{nn.mse.rm.1-1.#1.#3.00.0.2.raster.wmi.diagram}
		{nn.mse.rm.1-1.#1.#3.00.0.3.raster.diagram}
		{nn.mse.rm.1-1.#1.#3.00.0.3.raster.wmi.diagram}
		{2.1in}
		{A LUT weights diagram of a 1--1 NN for an image `#2',
		\(R^{r}_{a} = 0\), \(R^{r}_{b} = 0\).
		Connection \(x_{0} \to s(1, 0)\) (a) LUT component,
		(b) visit table,
		connection \(x_{1} \to s(1, 0)\) (c) LUT component,
		(d) visit table,
		connection \(s(1, 0) \to s(2, 0)\) (e) LUT component,
		(f) visit table.
		The \(i\) axes denote weight function LUT
		(a)(c)(e) or visit table (b)(d)(f) indices,
		\(x_{j}\) denotes \(j\)th input of the NN and
		\(s(l, n)\) denotes a node \(n\) in the \(l\)th layer.
		}
	\psfigureabc{nn.mse.rm.1-1.#1.#3.11.0.images}
		{nn.mse.rm.1-1.#1.#3.11.0.image.output.1000}
		{nn.mse.rm.1-1.#1.#3.11.0.image.output.10000}
		{nn.mse.rm.1-1.#1.#3.11.0.image.output.100000}
		{1.0in}
		{Images of the generalizing function after
		(a) \(1000\), (b) \(10000\) and (c) \(1\cdot 10^{5}\) iterations,
		respectively, of a
		1--1 NN for an image `#2', \(R^{r}_{a} = 1\cdot 10^{-4}\),
		\(R^{r}_{b} = 1\cdot 10^{-4}\).
		}
	\psfigureabcdef{nn.mse.rm.1-1.#1.#3.11.0.raster}
		{nn.mse.rm.1-1.#1.#3.11.0.1.raster.diagram}
		{nn.mse.rm.1-1.#1.#3.11.0.1.raster.wmi.diagram}
		{nn.mse.rm.1-1.#1.#3.11.0.2.raster.diagram}
		{nn.mse.rm.1-1.#1.#3.11.0.2.raster.wmi.diagram}
		{nn.mse.rm.1-1.#1.#3.11.0.3.raster.diagram}
		{nn.mse.rm.1-1.#1.#3.11.0.3.raster.wmi.diagram}
		{2.1in}
		{A LUT weights diagram of a 1--1 NN for an image `#2',
		\(R^{r}_{a} = 1\cdot 10^{-4}\),
		\(R^{r}_{b} = 1\cdot 10^{-4}\).
		Connection \(x_{0} \to s(1, 0)\) (a) LUT component,
		(b) visit table,
		connection \(x_{1} \to s(1, 0)\) (c) LUT component,
		(d) visit table,
		connection \(s(1, 0) \to s(2, 0)\) (e) LUT component,
		(f) visit table.
		The \(i\) axes denote weight function LUT
		(a)(c)(e) or visit table (b)(d)(f) indices,
		\(x_{j}\) denotes \(j\)th input of the NN and
		\(s(l, n)\) denotes a node \(n\) in the \(l\)th layer.
		}
}
\newcommand{\rasterdiagrams}[4] {
	\rasterdiagramsoutputweights{#1}{#2}{#3}{#4}
}
\newcommand{\psfigureoutputse}[4]{
        \begin{figure}[h!]
        \smallskip
        \begin{center}
	\begin{tabular}{cccc}
	        \epsfxsize=#3 \epsfbox{#2.output.1000.eps} \hspace{0.1in} &
        	\epsfxsize=#3 \epsfbox{#2.output.10000.eps} \hspace{0.1in} &
        	\epsfxsize=#3 \epsfbox{#2.output.100000.eps} \hspace{0.1in} &
        	\epsfxsize=#3 \epsfbox{#2.output.1000000.eps} \\[8pt]
        	\epsfxsize=#3 \epsfbox{#2.se.1000.eps} \hspace{0.1in} &
        	\epsfxsize=#3 \epsfbox{#2.se.10000.eps} \hspace{0.1in} &
        	\epsfxsize=#3 \epsfbox{#2.se.100000.eps} \hspace{0.1in} &
        	\epsfxsize=#3 \epsfbox{#2.se.1000000.eps} \\[10pt]
		(a) & (b) & (c) & (d) \\
	\end{tabular}
        \end{center}
        \caption{#4}
        \label{fig:#1}
        \end{figure}
}
\newcommand{\lineardiagrams}[6] {
	\psfigureoutputse{nn.mse.lm.#5.#1.#3.0.images}
		{nn.mse.lm.#5.#1.#3.0.image}
		{1.0in}
		{Output (top) and square error (bottom) images after
		(a) \(1000\), (b) \(10000\),
		(c) \(1\cdot 10^{5}\) and (d) \(1\cdot 10^{6}\) iterations,
		respectively, of a
		#6 NN for an image `#2'.
		}
	
        \begin{figure}[h!]
        \smallskip
        \begin{center}
	\input nn.mse.lm.#5.#1.#3.0.weights.table.tex
        \end{center}
        \caption{A weight diagram of a LWF #6 NN for an image `#2'.
		}
        \label{fig:nn.mse.lm.#5.#1.#3.0.weights.table}
        \end{figure}

}
\newcommand{\lineardiagramserror}[4] {
	
        \begin{figure}[h!]
        \smallskip
        \begin{center}
	\input comparison_linear.#1.tex
        \end{center}
        \caption{An MSE for an image `#2',
		\(\mu = #4\).
		}
        \label{fig:comparison_linear.#1}
        \end{figure}

}
\newcommand{\lineardiagramsnets}[4] {
	\lineardiagrams{#1}{#2}
		{#3}{#4}
		{4-4-1}{4-4-1}
	\lineardiagrams{#1}{#2}
		{#3}{#4}
		{8-8-1}{8-8-1}
	\lineardiagrams{#1}{#2}
		{#3}{#4}
		{16-1}{16-1}
	\lineardiagramserror{#1}{#2}
		{#3}{#4}
}
\subsection{DIFFUSION OF LUT WEIGHTS}
\label{sec:test_diffusion}
The training set generalized in this section, for visualization
purposes, is a raster image.
In each sample of the set there are three attributes
\((x,\,y,\,v)\). The attributes \(x\) and \(y\)
are the argument ones and represent coordinates in a two--dimensional
mesh and the \(v\) attribute is the value one.
The data set `circle' is seen in Fig.~\ref{fig:circle-data-set}(a).
\psfigureab{circle-data-set}
	{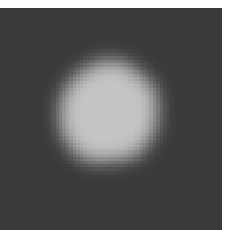}
	{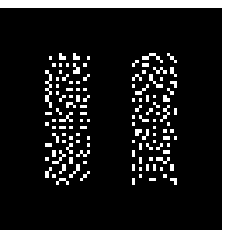}
	{1.2in}
	{The `circle' data set (a) and (b) its mask.}
The image has the resolution of \(64\times 64\).
The upper left corner pixel is at \((-0.5,\,-0.5)\) and
the lower right corner pixel is at \((0.5,\,0.5)\).
Black pixels on the images are denoted by \(-0.5\) and
white ones by \(0.5\), with a gray scale between the
two values. The NN generalization function will be shown in a similar way,
but the values less than
\(-0.5\) or greater than \(0.5\) will also be shown as black or white pixels,
respectively. The mask in Fig.~\ref{fig:circle-data-set}(b) shows
by white pixels the respective samples that are chosen for the training.

Let us first test the introduced NNs without the diffusion of the weight
functions,
to compare it later with NNs that have the diffusion.
To disable the weight functions diffusion, let \(R^{r}_{b} = 0\).
Let the
LUT smoothing will also be disabled by setting \(R^{r}_{a} = 0\)
to show the generalization similar to that seen in Fig.\ref{fig:wftypes}(c).
Let the \(\zeta\) coefficient be equal to \(1\) in the examples in
this section, as in the section not the time efficiency is tested,
but the generalization ability is presented, and \(\zeta = 1\) allows
for smoother diagrams of weights.
In Fig.~\ref{fig:nn.mse.rm.1-1.circle.s3.l1.00.0.images}
images of the generalizing function representing the approximated
training set
are shown for a 1--1 NN at some different iterations.
The used training set leaves relatively large regions
in the space of the argument attributes
unknown by a trained NN. Because of the generalization ability the NN should
`extrapolate' learned samples over these regions.
As can be seen in
Fig.~\ref{fig:nn.mse.rm.1-1.circle.s3.l1.00.0.images}, the NN without diffusion
generalizes relatively poorly at the \(1\cdot 10^{5}\)th iteration,
showing problems resulting from the lack of both smoothness
like it was described in \cite{piazza1993lut}.
In Fig.~\ref{fig:nn.mse.rm.1-1.circle.s3.l1.00.0.raster},
diagrams showing the nonlinear weight functions
and the visit tables at some iterations of the training process are
presented. The intensity representing the values in visit tables is
nonlinearly related to the values, to make the smaller ones better visible.
The lack of diffusion of the LUT weight functions is clearly seen.

Let us then test an identical NN, but with the standard values of
\(R^{r}_{a}\) and \(R^{r}_{b}\) of \(1\cdot 10^{-4}\).
In Fig.~\ref{fig:nn.mse.rm.1-1.circle.s3.l1.11.0.images}
a much improved generalization can be seen
after the first \(1\cdot 10^{5}\) iterations,
in compare to that shown in
Fig.~\ref{fig:nn.mse.rm.1-1.circle.s3.l1.00.0.images}.
In Fig.~\ref{fig:nn.mse.rm.1-1.circle.s3.l1.11.0.raster}
the diffusion of LUT weight functions can be seen.

Some tests with various values of the diffusion speed coefficient
\(R^{r}_{b}\) will also be performed in Sec.~\ref{sec:sparse_data_sets}.
\rasterdiagrams{circle}{circle}
	{s3.l1}{\(s_{\mathrm{0}}\)}

\subsection{SMALL SIZE DATA SETS}
\label{sec:tests-small-size}
The type of neural networks presented in this paper were designed for
generalization of sets of a very high complexity or highly nonlinear,
for example needing hundreds of thousands of samples to be roughly
represented, or that have patterns like the `two spirals' set
\cite{lang1988spirals}, tested in the next section. Many generalized data
sets, however, are much smaller and more linear.
Yet, it still may be a difficult task to
generalize them well -- in \cite{draghici2001constraint} performance
comparison of several neural-- and non--neural learning algorithms shows
substantial variance in the percentage of properly classified test
samples in the case of some data sets from the UCI repository \cite{blake1998uci}, of which
all have less than one thousand samples. It can be important for a
learning machine to have a good performance on a wide range of data sets, so
in this section performance comparison is performed on some relatively small
data sets. In this test, classification results of classic neural networks with linear
weight functions, several other neural and non--neural learning algorithms,
and the introduced neural networks are compared. The generalized sets are from the
mentioned UCI
repository and generally have patterns of a moderate nonlinearity.

The coefficient \(R^{r}_{b}\) was set to \(0.02\) to increase the
diffusion rate so to counterpart the relatively sparse samples
in the generalized sets.

The generalized sets were randomly divided into
training sets containing 80\% of samples and test sets containing
20\% of samples. 10 runs of each tested architecture of the classic LW networks, and of the
introduced NLW networks, were performed, and the average values were shown.
For comparison, the SVM \cite{vapnik1995, vapnik1998} machines of the
type \(\nu\)--SVM \cite{schoelkopf2000new, schoelkopf2001estimating}
with radial basis function kernel were also tested. Because of their relatively high
training speed in the case of the small data sets, SVMs were run with ten--fold crossvalidation
of the \(c\) and \(\gamma\) coefficients.
To test the SVMs, the LIBSVM
package \cite{chih2001libsvm} was used. The classification results, averaged over the tested sets,
for some other learning machines:
C4.5 using classification rules \cite{quinlan1993}, incremental decision tree
induction ITI \cite{utgoff1989, utgoff1997}, linear machine decision tree
LMDT \cite{utgoff1991}, learning vector quantization LVQ \cite{kohonen1988lvq,
kohonen1995lvq}, induction of oblique trees OCI \cite{heath1993}, Nevada backpropagation
NEVP based on Quickprop \cite{fahlman1988faster}, \(k\)--nearest neighbors
with \(k = 5\) K5, Q* and radial basis functions RBF \cite{poggio1990, musavi1992}
were computed using results from tests in \cite{draghici2001constraint}.
A very good comparison of the algorithms can be found in \cite{eklund2000comparative}.

The training limit for the NNs was 10000 iterations.
The tested LW networks had considerably more connections to make
their single training iteration similarly fast to that of the tested NLW network.
In Table \ref{tab:low-complexity-classification-nn} results are shown for
the classic feedforward networks, the diffused feedforward networks
and for \(\nu\)--SVM with ten--fold crossvalidation of \(c\) and \(\gamma\)
coefficients.
The letter X symbolizes the number of inputs, equal to the number of argument
attributes in samples in a given set.
The networks LW X-16-16-1 and NLW X-8-8-1 had similar time complexity,
but the latter one performed better on average.
Table \ref{tab:low-complexity-classification} shows
the average results for the same data sets,
except for RBF neural networks that have the result for the `Zoo' set missing,
 reported in \cite{draghici2001constraint},
for various other learning machines. In \cite{draghici2001constraint} detailed results
for individual data sets can be found.
The comparisons should be interpreted cautiously --
the division into the
training and test sets may be different, giving various classification
results -- within the learning machines LW, NLW and SVM tested by
the author it was the same, though. It can be seen that SVM had best average results,
and NLW was the second in all of the tests.
It can also be seen, that the 
results of NLW networks are relatively similar for very different number of connections and
two different \(r_{\mathrm{res}}\) values.
\begin{table}[h!]
\begin{center}
{\tiny
\begin{tabular}{r|p{0.45in}@{ }p{0.65in}@{ }p{0.65in}@{ }p{0.4in}@{ }p{0.5in}@{ }p{0.6in}@{ }p{0.7in}}
Data set &		LW X-8-1	& LW X-16-16-1	& LW X-32-32-1	& \(\nu\)--SVM & NLW X-4-1\newline\(r_{\mathrm{res}}=16\)	& NLW X-8-8-1\newline\(r_{\mathrm{res}}=16\)	& NLW X-16-16-1\newline\(r_{\mathrm{res}}=64\) \\[6pt]
\hline
\hline\\[-3pt]
Glass	&		67.68		&	66.74  &	70.93				&	74.42	&	76.74	& 76.05			& \textbf{80.00}		 \\[3pt]
\hline\\[-4pt]
Ionosphere &	94.71		&	95.00  &	\textbf{95.43}		&	92.86	&	91.86	& 93.29			& 85.89		 \\[3pt]
\hline\\[-4pt]
Wine &			\textbf{98.61}		&	95.28  &	97.22		&	97.22	&	96.39	& 94.72			& 95.28		 \\[3pt]
\hline\\[-4pt]
Pima &			77.53		&	77.53  &	76.56				&	77.27	&	\textbf{78.12}	& 75.91			& 75.07		 \\[3pt]
\hline\\[-4pt]
Bupa &			60.72		&	59.86  &	63.63				&	\textbf{69.57}	&	60.87	& 62.03			& 66.09		 \\[3pt]
\hline\\[-4pt]
Tic tac toe &	96.93		&	96.93  &	97.04		&	\textbf{100.00}	&	96.78	& 96.20			& 97.04		 \\[3pt]
\hline\\[-4pt]
Balance &		90.64		&	91.44  &	90.72				&	\textbf{100.00}	&	96.16	& 96.48	& 95.68		 \\[3pt]
\hline\\[-4pt]
Iris &			96.00		&	95.33  &	95.33		&	\textbf{96.67}	&	94.67	& 95.33			& 94.66		 \\[3pt]
\hline\\[-4pt]
Zoo &			86.00		&	\textbf{90.50}  &	88.00		&	85.00	&	86.50	& 88.50			& 87.50		 \\[3pt]
\hline
\hline\\[-1pt]
 Average &		85.42		&	85.40  &	86.10				&	\textbf{88.11}	&	86.45	& 86.50			& 86.36		 \\[3pt]
\end{tabular}
}
\end{center}
\caption{Comparison of classification results of small size data sets for several LW and
NLW networks and \(\nu\)--SVM, in percents.}
\label{tab:low-complexity-classification-nn}
\end{table}
\begin{table}[h!]
\begin{center}
{\tiny
\begin{tabular}{r|r@{ }r@{ }r@{ }r@{ }r@{ }r@{ }r@{ }r@{ }r@{ }r@{ }r@{ }r@{ }r@{ }r@{ }r@{ }r}
		&		 C4.5 &	C4.5r &	  ITI &	LMDT &	  CN2 &	  LVQ &	  OC1 &	 NEVP &	  K5 &	   Q* &	RBF &	CBD \\[6pt]
\hline
\hline\\[-3pt]
 Average &	79.89 &	82.71 &	82.25 &	\textbf{84.76} &	83.52 &	76.97 &	75.61 &	82.30 &	76.68 &	74.52 &	75.29 &	81.94 \\[3pt]
\end{tabular}
}
\end{center}
\caption{Comparison of average classification results of small size data sets, the same as in
Table \ref{tab:low-complexity-classification-nn}, in percents, for several
neural and non--neural learning machines.}
\label{tab:low-complexity-classification}
\end{table}

\subsection{TWO SPIRALS}
\label{sec:sparse_data_sets}
%

Because the data sets tested in this section are relatively sparse
as for the introduced networks, the resolution of the LUT tables was decreased
to \(r_{\mathrm{res}} = 16\) and, like it was done in tests in
Sec,~\ref{sec:tests-small-size}, the diffusion was `strengthened'
by using relatively large \(R^{r}_{b}\) values.
The NLW networks are compared to the classic ones and to some other
learning machines.

\begin{figure}[h!]
        \vskip 0.0in
        \begin{center}
        \begin{tabular}{cc}
			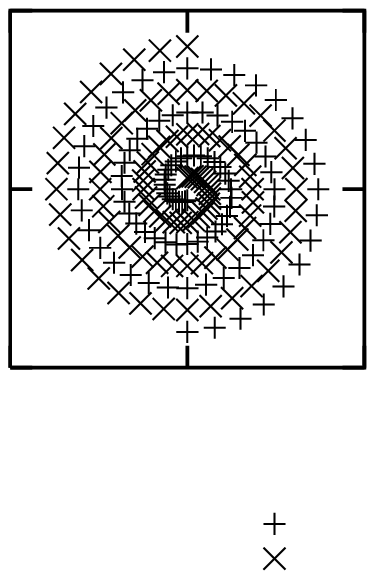 &
			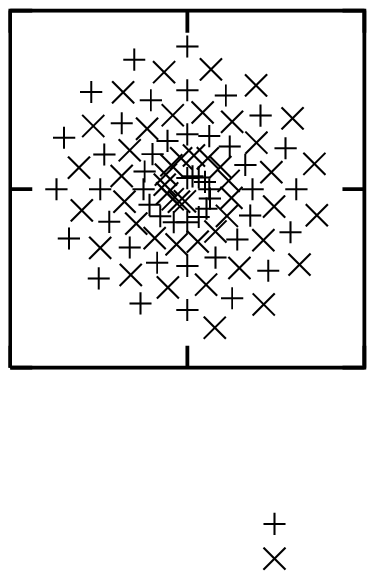 \\
			(a) & (b) \\
		\end{tabular}
        \end{center}
        \caption{The (a) `two spirals' and (b) `two spirals sparse' data sets.}
        \label{fig:spirals}
        \end{figure}

Let us first test the generalization
of the `two spirals' set, one of the standard benchmarks for learning
machines \cite{lang1988spirals}. This set, after centering around
the point \((0, 0)\), is shown in Fig.~\ref{fig:spirals}(a).

Each sample in the set has three attributes \((x_{0}, x_{1}, y)\),
the first two being the argument attributes and the last one the value
attribute.
Even though the spirals may be regarded as relatively well defined because of
the density of the samples determining them, this set is known to be
a very hard two--class problem to learn by classic feedforward neural networks
using the error backpropagation family of learning algorithms.
In \cite{lang1988spirals} it was reported that the task could not be solved with the
tested classic feedforward networks with connections only between
neighboring layers, so to classify each sample in the training set, 
a special architecture was developed, where each node was connected to all nodes
in the subsequent layers and the network was trained using error backpropagation
with momentum. Several other trials have been undertaken for improving the learning algorithms
to train feedforward neural networks more efficiently. For example in \cite{fahlman1990cascade}
a learning algorithm is developed that grows the trained neural network by adding
new trained units to the network. The algorithm was successfully applied to the
two spirals problem, but even though the trained neural network learned to classify
all samples in the training set, its generalization quality was
relatively poor -- 
the decision border was very rough and it even crossed the
arms of the spirals in some places.
Images of the generalization function of the network can be found also in
\cite{fahlman1990cascade}.
The radial basis function neural networks \cite{moody1989fast}
even with the advanced techniques
like dynamic decay adjustments \cite{berthold1995boosting}, show the problem of the
lack of a `long range' generalization -- as can be seen in the images in
\cite{berthold1995boosting}, the samples are classified correctly,
but the regions far from the samples seem to have little or nothing in
common to the positive or negative values of the individual samples. The
neural networks presented in \cite{perwass2003spherical}, that have neurons
whose decision borders are hyperspheres, 
have the problem with the `long range' generalization as well,
as can be seen in the images in \cite{perwass2003spherical}. 


Let the generalization functions of the networks be sampled, and presented as
two--dimensional gray scale raster images
of the size \(64 \times 64\), in the same manner as was done
in Sec.\ref{sec:test_diffusion}.
In Fig.~\ref{fig:training-spirals-2} generalization functions are
shown for NLW networks trained at two different values of
the diffusion speed coefficient \(R^{r}_{b}\).
Fig.~\ref{fig:spirals-2-binary} shows
classification results for these networks and for a \(\nu\)--SVM
\cite{schoelkopf2000new, schoelkopf2001estimating}
The results for the tested LW networks are not shown as
they were not able to even classify
the training set within \(1\cdot 10^{6}\) iterations. The SVM performed very
good on the set. The NLW network
with the diffusion speed coefficient \(R^{r}_{b} = 1\cdot10^{-4}\)
generalized the training set with a somewhat rough decision border.
Increasing the diffusion speed by increasing the value of the
diffusion speed coefficient \(R^{r}_{b}\) to \(0.01\)
caused that the generalization was much better.
In \cite{solazzi2000multidimensional} a neural network is introduced
with adaptive multidimensional spline activation functions, and is shown that the network
has excellent results of the generalization results of the
two spirals set, similar to these in Fig.~\ref{fig:spirals-2-binary}(a) and (c),
if the adaptive spline activation function is two--dimensional. Yet the `two spirals' set
has a generalization function that is also two--dimensional.
If the adaptive activation functions have a single dimension, like in the networks
presented by 
	Uncini, Capparelli and Piazza \cite{uncini1998},
	Vecci, Piazza and Uncini \cite{vecci1998learning}, and
	Guarnieri and Piazza \cite{guarnieri1999},
so that the `scale up'
of the dimension by the high dimension regressor is non--zero for
the `two spirals' set, the images demonstrated in
\cite{solazzi2000multidimensional} shows substantial artifacts,
much larger than these in Fig.~\ref{fig:spirals-2-binary}(b).
\newcommand{\pstablespiraliteration}[1]{
	\begin{tabular}{r}#1\\[33pt]\end{tabular}\hspace{-0.2in}
}
\newcommand{\pstablespiralitem}[2]{
	\epsfysize=0.5in \epsfbox{mse.#1.image.output.#2.eps}
}
	\begin{figure}[h!]
	\smallskip
	\begin{center}
		\begin{tabular}{lccccc}
			Iteration\hspace{0.0in} &
			\begin{minipage}{0.6in}\scriptsize
				NLW\\
				\mbox{2-32-32-1}\\ \(R^{r}_{b} =\)\\ \(1\cdot 10^{-4}\)\\
			\end{minipage} &
			\begin{minipage}{0.6in}\scriptsize
				NLW\\
				\mbox{2-32-32-1}\\ \(R^{r}_{b} = 0.01\)\\ ~\\
			\end{minipage}
			\\
			\pstablespiraliteration{\(1\cdot 10^{4}\)} &
				\pstablespiralitem{spirals-2-r16-8-r}{10000} &
				\pstablespiralitem{spirals-2-r16-8-r-33}{10000} \\[-10pt]
			\pstablespiraliteration{\(1\cdot 10^{5}\)} &
				\pstablespiralitem{spirals-2-r16-8-r}{100000} &
				\pstablespiralitem{spirals-2-r16-8-r-33}{100000} \\[-10pt]
			\pstablespiraliteration{\(1\cdot 10^{6}\)} &
				\pstablespiralitem{spirals-2-r16-8-r}{1000000} &
				\pstablespiralitem{spirals-2-r16-8-r-33}{1000000} \\[-10pt]
		\end{tabular}
	\end{center}
	\caption{Images of the generalizing function for the `two spirals' set.}
	\label{fig:training-spirals-2}
	\end{figure}
\psfigureabc{spirals-2-binary}
	{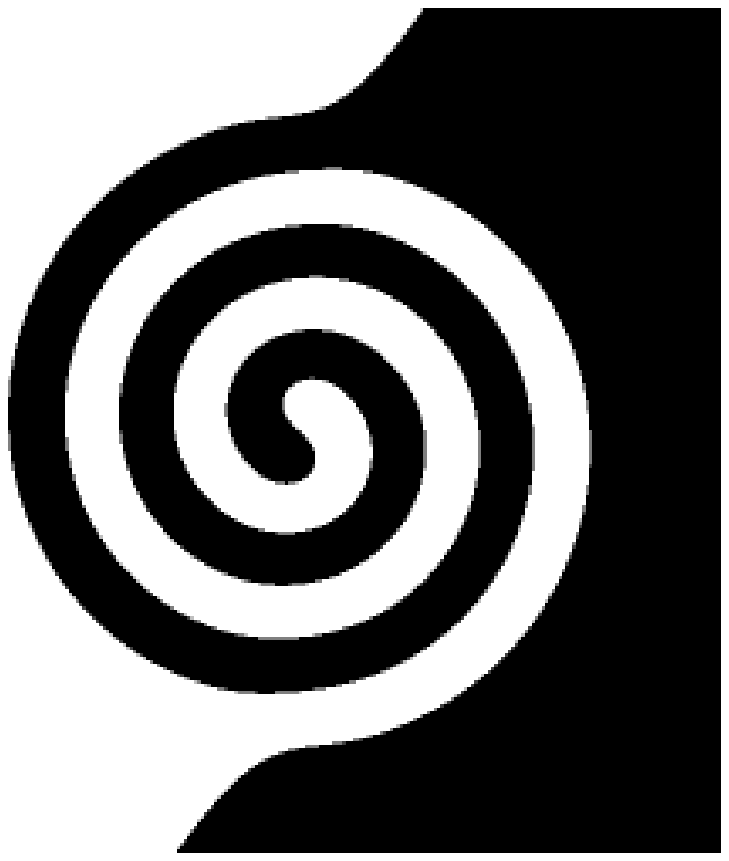}
	{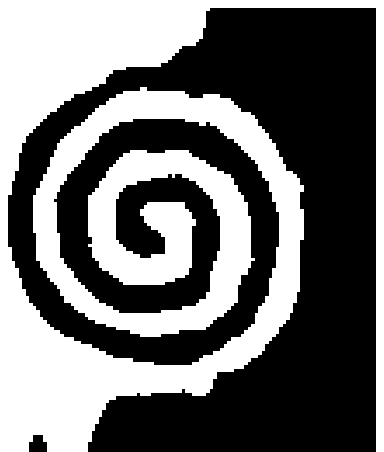}
	{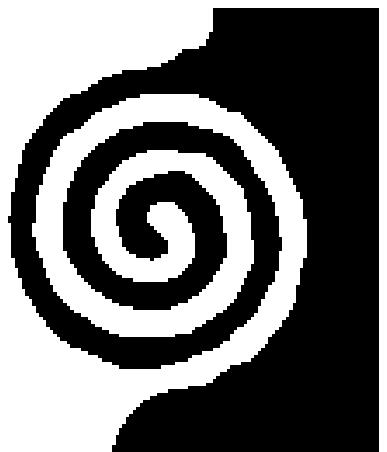}
	{1in}
	{Classification of the `two spirals' set by
		(a) SVM with radial basis kernel function, \(c = 100, \gamma = 150\),
		and NLW 2--32--32--1 at the \(1\cdot 10^{6}\)th iteration,
		at (b) \(R^{r}_{b} = 1\cdot 10^{-4}\),
		(c) \(R^{r}_{b} = 0.01\).
	}

\begin{figure}[h!]
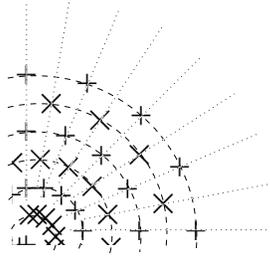

        \smallskip
        \begin{center}
	\input spirals-patterns-2.pstex_t
        \end{center}
        \caption{Two families of patterns in the `two spirals sparse' data set,
	 denotes with different types of lines.}
        \label{fig:spirals-patterns-2}
        \end{figure}

Let us now discuss another training set, derived from the previous
one.
Let the samples within each of the spiral arms be counted
from the inner beginning of each arm.
This set is created by removing each odd sample in one of the
spiral arms and each even sample in the other arm.
Let the set be called `two spirals sparse'. This set is shown
in Fig.~\ref{fig:spirals}(b). Such a way of removing the
samples was used to obtain a special type of patterns in the
set. The patterns create two families, `along arms' and `radial',
as illustrated in Fig.~\ref{fig:spirals-patterns-2}.

It can be said that in the inner side
of the spirals the `along arms' pattern is stronger than
the `radial' one, because of the relationship between appropriate distances
between samples. Conversely, the `radial' pattern is stronger
in the outer regions of the spiral arms.
The third type of pattern is created by the outer parts
of the spirals. Because the halves are not covered from
the outside by any samples, the value attributes of the
samples creating the halves may possibly be extrapolated
to the outside, so that outside the spirals the generalizing
function may roughly have values greater than \(0\) for \(x_{1} > 0\)
and less than \(0\) for \(x_{1} < 0\).
Let the task be to generalize the discussed set so that the strengths
of the two families of patterns and of the third discussed pattern
would be appropriately reflected in the generalizing function.
The gradual transition of patterns in the discussed set will allow
for testing the evenness of generalizing different regions of the
space of argument attributes of the samples.

In Fig.~\ref{fig:training-spirals} generalization functions
at some different iterations for NLW networks trained
at two different values of the diffusion speed \(R^{r}_{b}\) are shown.
Fig.~\ref{fig:spirals-binary} shows
the classification results for \(\nu\)--SVM and NLW networks at two
different \(R^{r}_{b}\) values.
	\begin{figure}[h!]
	\smallskip
	\begin{center}
		\begin{tabular}{lcccccc}
			Iteration\hspace{0.0in} &
			\begin{minipage}{0.6in}\scriptsize
				NLW\\
				\mbox{2-32-32-1}\\ \(R^{r}_{b} =\)\\ \(1\cdot10^{-4}\)\\
			\end{minipage} &
			\begin{minipage}{0.6in}\scriptsize
				NLW\\
				\mbox{2-32-32-1}\\ \(R^{r}_{b} = 0.01\)\\ ~\\
			\end{minipage} &
			\begin{minipage}{0.6in}\scriptsize
				NLW\\
				\mbox{2-32-32-1}\\ \(R^{r}_{b} = 0.1\)\\ ~\\
			\end{minipage}
			\\
			\pstablespiraliteration{\(1\cdot 10^{4}\)} &
				\pstablespiralitem{spirals-r16-8-r}{10000} &
				\pstablespiralitem{spirals-r16-8-r-33}{10000} &
				\pstablespiralitem{spirals-r16-8-r-44}{10000} \\[-10pt]
			\pstablespiraliteration{\(1\cdot 10^{5}\)} &
				\pstablespiralitem{spirals-r16-8-r}{100000} &
				\pstablespiralitem{spirals-r16-8-r-33}{100000} &
				\pstablespiralitem{spirals-r16-8-r-44}{100000} \\[-10pt]
			\pstablespiraliteration{\(1\cdot 10^{6}\)} &
				\pstablespiralitem{spirals-r16-8-r}{1000000} &
				\pstablespiralitem{spirals-r16-8-r-33}{1000000} &
				\pstablespiralitem{spirals-r16-8-r-44}{1000000} \\[-10pt]
		\end{tabular}
	\end{center}
	\caption{Images of the generalizing function for the `two spirals sparse' set.}
	\label{fig:training-spirals}
	\end{figure}
\psfigureabcdefhoriz{spirals-binary}
	{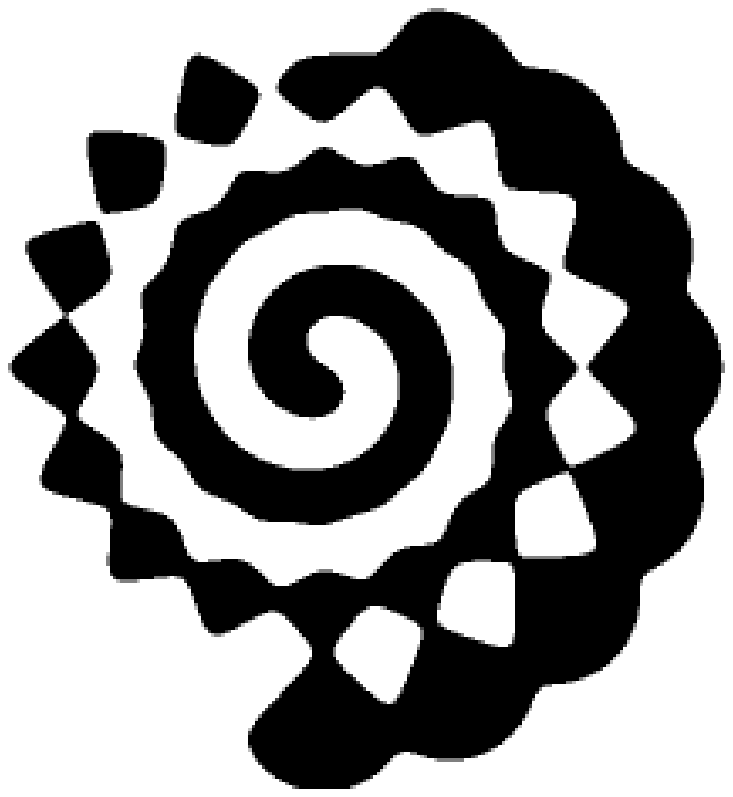}
	{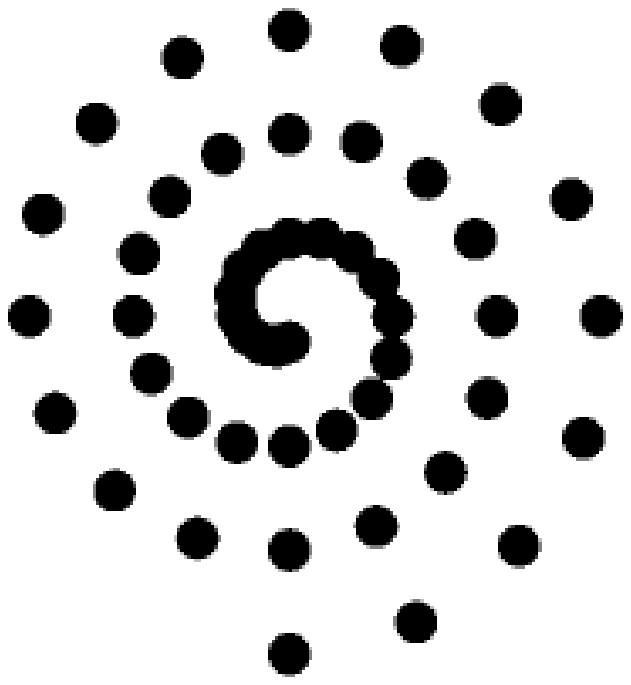}
	{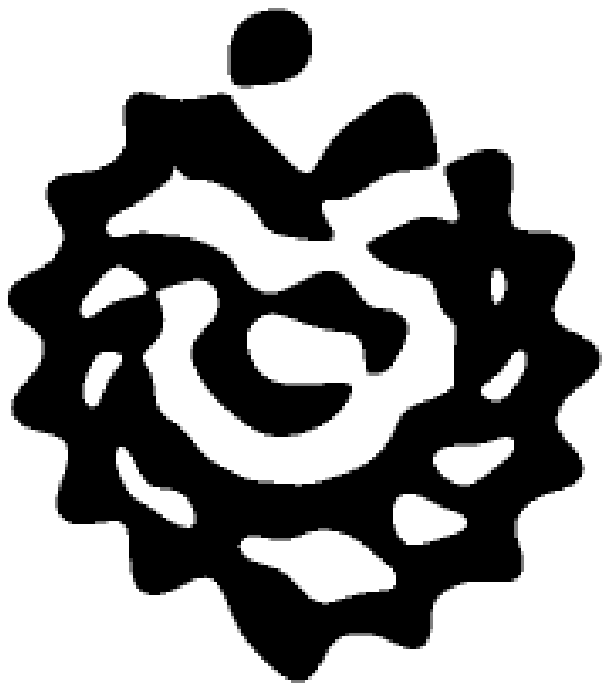}
	{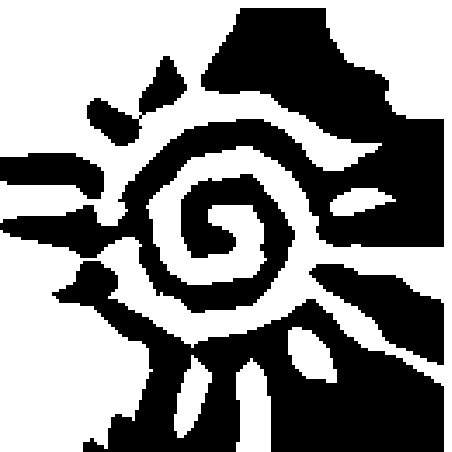}
	{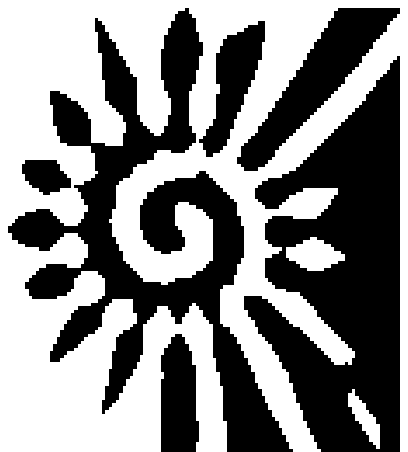}
	{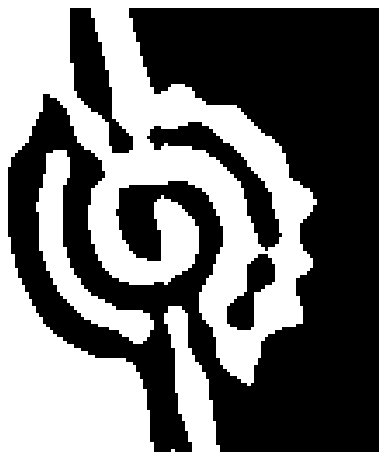}
	{1in}
	{Classification of the `two spirals sparse' set by
		(a) \(\nu\)--SVM with radial basis kernel, \(c = 1, \gamma = 500\),
		(b) \(\nu\)--SVM with radial basis kernel, \(c = 1, \gamma = 10000\),
		(c) one class SVM with radial basis kernel, \(c = 100, \gamma = 200\), 
		and NLW 2--32--32--1 at the \(1\cdot 10^{6}\)th iteration,
		at (d) \(R^{r}_{b} = 1\cdot10^{-4}\),
		(e) \(R^{r}_{b} = 0.01\),
		(f) \(R^{r}_{b} = 0.1\).
	}
The tested LW networks did not classify
the training set within \(1\cdot 10^{6}\) iterations.
The NLW network
with diffusion speed coefficient \(R^{r}_{b} = 1\cdot10^{-4}\)
generalized the training set but with rather severe problems.
After increasing values of the
diffusion speed coefficient to \(R^{r}_{b} = 0.01\)
the generalization was much better -- all three discussed types
of patterns are seen. Further increase of the diffusion
speed coefficient \(R^{r}_{b}\) to \(0.1\) rather did not
give any improvements -- as can be seen in the generalizing functions
in Fig.\ref{fig:training-spirals},
the learning process slowed down,
and at the \(1\cdot 10^{6}\)th iteration, the generalizing functions
seemed to be relatively less `even'.
The SVM with the settings like in the previous test with the `two spirals'
set, substantially underfitted the `two spirals sparse' set, so
it was with tested various different \(\gamma\) values, yet none
of the tested SVMs shown results so fine as the NLW networks --
SVMs tended to give asymmetric decision borders and to neglect
the `radial' pattern, and the one class SVM
\cite{schoelkopf2001estimating}
gave particularly spurious results.

\subsection{LEARNING A LARGE SIZE SET}
In this section, the `storage capacity' of a trained neural network
against time is tested, that is the ability to memorize the features
in a set, and the speed of the memorizing.
Generalization of sets like `two spirals' \cite{lang1988spirals}
shows that even large, and thus slow LW networks might have serious
problems with just the memorizing and the speed of the memorizing.
The training set tested in this section
is relatively complex and, to test the high dimension
regressor, the samples have five input attributes each.
Standard values of coefficients were used for training the neural
networks with this complicated set. 

Because the author could not find a standard benchmark of the required
complexity and number of samples, custom data set `md-2' was used.
Let the generating function of the set be as follows:
\begin{equation}
	\begin{array}{rcl}
	y\left(x_{0}, x_{1}, x_{2}, x_{3}, x_{4}\right) & = &
		\sin(4x_{0})
		\cos(2x_{1} + 3x_{2}) \\
		& & \left(\left(\frac{\sin\left(10x_{2} + 10x_{3}\right) + 1}{2}\right)
			^{1/2} - 1/2\right) \\
		& & \sin(x_{3} - 4x_{1}x_{4}) \\
		& & \left(\left(\frac{\sin(10x_{0} - 10x_{2} + 10x_{3}) + 1}{2}\right)
			^{1/2} - 1/2\right) \\
		& & \cos(5x_{1}x_{2}x_{4})
		.
	\end{array}
\label{eq:md-2}
\end{equation}
Using this equation, tuples \((x_{0}, x_{1}, x_{2}, x_{3}, x_{4}, y)\)
were generated, where \(x_{i}\) were random values
\begin{equation}
	x_{i} = \mathrm{rand}() - 0.5 \quad i = 0, 1, \ldots 4,
\end{equation}
where \(\mathrm{random}()\) is a uniform random number generator,
generating values in the range \(\left<0, 1\right)\).
\(1.8\cdot 10^{6}\) such tuples were used in the training set, and \(2\cdot 10^{5}\)
independently generated tuples were used in the test set. The data set
intentionally does not contain noise and is quite densely represented, to
test the discussed `storage capacity'.

\begin{figure}[h!]
	\begin{center}
		\hspace*{-1.0in}
		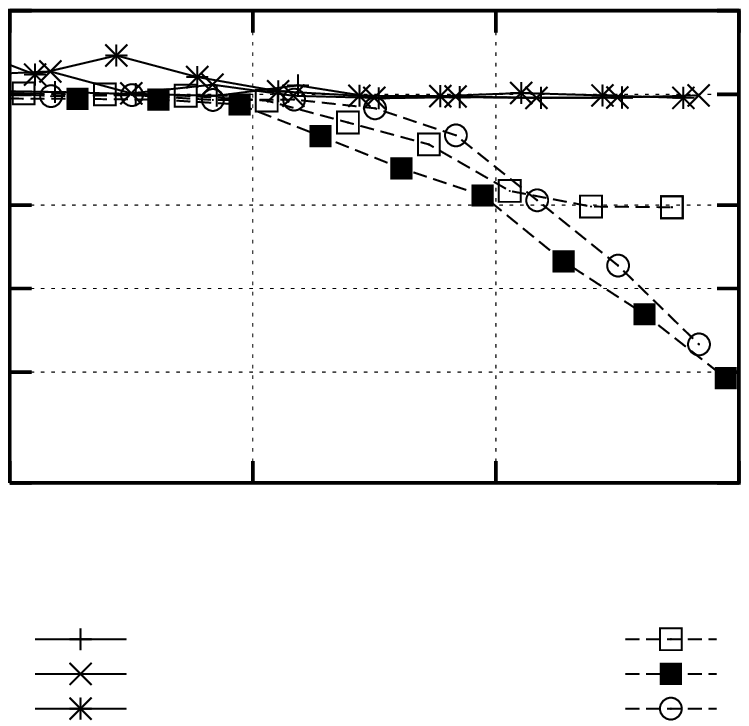
	\end{center}
	\caption{MSE diagram for different neural networks trained with the `md-2' set,
		against estimated time.}
	\label{fig:mse.md-2.diagram.tex}
\end{figure}

Fig.~\ref{fig:mse.md-2.diagram.tex} shows a diagram of mean square error,
denoted by MSE, against
estimated times for several
LW and NLW networks, trained with the set `md-2'.
It is seen that the tested NLW networks
reach MSE even about ten times lower after similar training times
in comparison to the tested LW networks.

\section{CONCLUSIONS}
\label{sec:conclusions}
The neural networks with diffused weight functions and with the property of
selective parameters of the functions showed good performance over
a wide range of tested data sets. In particular, they performed very good, in 
compare to the classic neural networks ant to the tested SVMs, in the case of
the subtle patterns in the `two spirals sparse' and `camomiles--m' sets.

\bibliography{nn}
\bibliographystyle{plainnat}

\end{document}